\documentclass{article}

\usepackage[numbers]{natbib}
\usepackage[final]{nips_2017}
\usepackage{subcaption}
\usepackage{graphicx}
\usepackage{hyperref}
 \usepackage[final]{nips_2017}
\usepackage[flushleft]{threeparttable} 
\usepackage{booktabs,caption}
\usepackage[utf8]{inputenc} 
\usepackage[T1]{fontenc}    
\usepackage{hyperref}       
\usepackage{url}            
\usepackage{booktabs}       
\usepackage{amsfonts}       
\usepackage{nicefrac}       
\usepackage{microtype}      

\usepackage{graphicx}
\usepackage{amssymb,amsmath,bm}
\usepackage{textcomp}
\usepackage{multirow}
\usepackage{dsfont}
\usepackage{color}
\usepackage[normalem]{ulem}
\usepackage{rotating}
\usepackage{makecell}
\usepackage{comment}
\usepackage{amsfonts,amssymb}
\usepackage{tikz}
\usepackage{verbatim}
\usepackage{tabularx}
\usepackage{subcaption}

\title{A Benchmarking Environment for Reinforcement Learning Based Task Oriented Dialogue Management}

\author{
  I\~nigo~Casanueva$^*$, Pawe\l~Budzianowski$^*$, Pei-Hao~Su$^*$,\\
  \textbf{Nikola~Mrk{\v s}i{\' c}, Tsung-Hsien~Wen, Stefan~Ultes, Lina~Rojas-Barahona, }  \\  \textbf{Steve Young and Milica~Ga{\v s}i{\' c}}  \\
  Department of Engineering\\
  University of Cambridge\\
  \texttt{\{ic340,pfb30,phs26\}@cam.ac.uk}\\
}

\begin{document}

\maketitle
{\let\thefootnote\relax\footnotetext{* equal contribution.}}
\vspace{-1em}
\begin{abstract}
\vspace*{-0.3cm}
Dialogue assistants are rapidly becoming an indispensable daily aid. To avoid the significant effort needed to hand-craft the required dialogue flow, the Dialogue Management (DM) module can be cast as a continuous Markov Decision Process (MDP) and trained through Reinforcement Learning (RL). Several RL models have been investigated over recent years. However, the lack of a common benchmarking framework makes it difficult to perform a fair comparison between different models and their capability to generalise to different environments. Therefore, this paper proposes a set of challenging simulated environments for dialogue model development and evaluation.  To provide some baselines, we investigate a number of representative parametric algorithms, namely deep reinforcement learning algorithms - DQN, A2C and Natural Actor-Critic  and compare them to a non-parametric model, GP-SARSA. Both the environments and policy models are implemented using the publicly available PyDial toolkit and released on-line, in order to establish a testbed framework for further experiments and to facilitate experimental reproducibility.
\end{abstract}

\section{Introduction}
\begin{figure}[!t]
  \centering
 \includegraphics[width=0.8\textwidth,trim={4cm 8.6cm 4cm 5.5cm},clip]{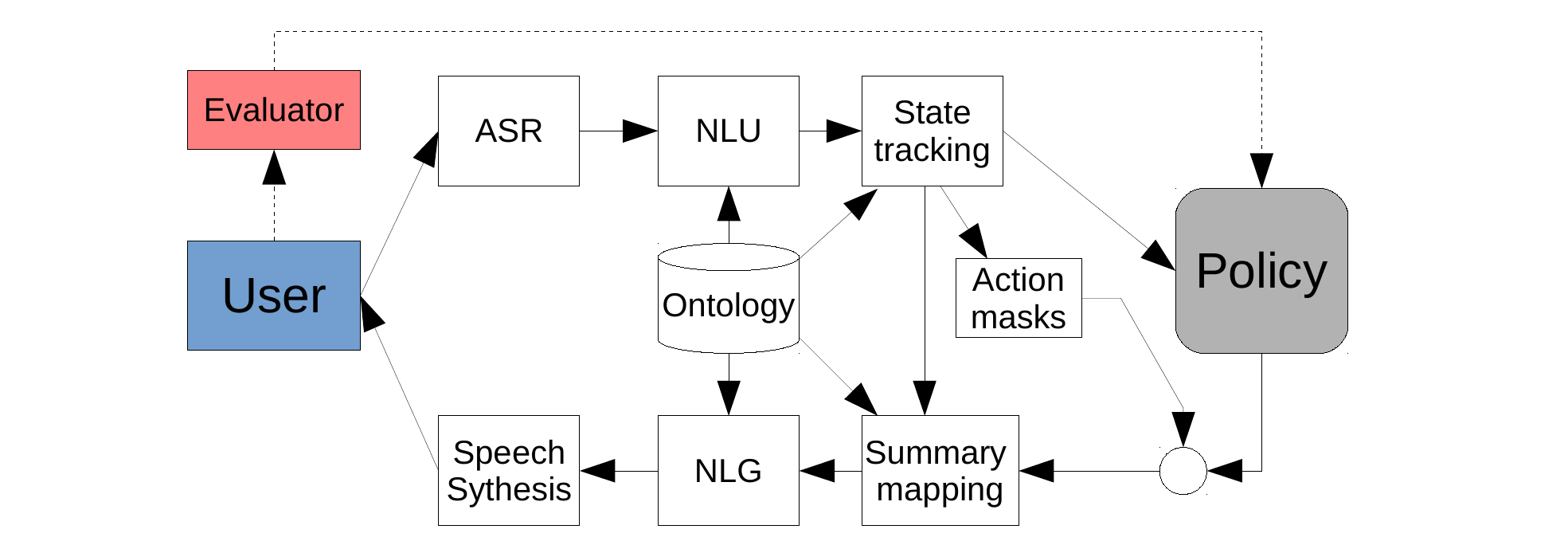}
  \caption{Spoken dialogue system environment used in this benchmark.}
  \label{fig:rl_env}
  \vspace*{-0.5cm}
\end{figure}

In recent years, due to the large improvements achieved in Automatic Speech Recognition (ASR), Natural Language Understanding (NLU) and machine learning techniques, dialogue systems have gained much attention in both academia and industry. Two directions have been intensively researched: open-domain chat-based systems \citep{vinyals2015neural,serban2016building} and task-oriented dialogue systems \citep{POMDP_williams,young2013pomdp}. The former cover non-goal driven dialogues about general topics. The latter aim to assist users to achieve specific goals via natural language, making it a very attractive interface for small electronic devices. Under a speech-driven scenario, Spoken Dialogue Systems (SDSs) are typically based on a modular architecture (Fig. \ref{fig:rl_env}), consisting of input processing modules (ASR and NLU modules), Dialogue Management (DM) modules (belief state tracking and policy) and output processing modules (Natural Language Generation (NLG) and speech synthesis). The domain of a SDS is defined by the ontology, a structured representation of the database of the system defining the \textit{requestable slots}, \textit{informable slots} and database entries (i.e. the type of entities users can interact with and their properties). Part of the dialogue flow in such an architecture is explained schematically in Figure \ref{fig:dial_flow} in Appendix \ref{apx:dial_flow}.


The DM module is the core component of a modular SDS, controlling the conversational flow of the dialogue. Traditional approaches have been mostly based on handcrafted decision trees covering all possible dialogues outcomes. However, this approach does not scale to larger domains and it is not resilient to noisy inputs resulting from ASR or NLU errors. Therefore, data-driven methods have been proposed to learn the policy automatically, either from a corpus of dialogues or from direct interaction with human users \citep{wen2016network,gasic2014gaussian}. 

Supervised learning can be used to learn a dialogue policy, training the policy model to "mimic" the responses observed in the training corpora \citep{wen2016network}. This approach, however, has several shortcomings. In a spoken dialogue scenario, the training corpora can not be guaranteed to represent optimal behaviour. The effect of selecting an action on the future course of the dialogue is not considered and this may result in sub-optimal behaviour. 
In addition, due to the large size of the dialogue state space,
the training dataset may lack sufficient coverage. 

To tackle the issues mentioned above, this task is frequently formulated as a planning (control) problem \cite{young2013pomdp}, solved using Reinforcement Learning (RL) \cite{sutton1999between}. In this framework, the system learns by a trial-and-error process governed by a potentially delayed reward signal. Therefore, the DM module learns to plan actions in order to maximise the final outcome. Recent advances such as Gaussian Process (GP) based RL \citep{gasic2014gaussian,casanueva2015knowledge} and deep RL methods \citep{mnih2013playing,silver2016mastering} have led to  significant progress in data-driven dialogue modelling, showing that general algorithms such as policy gradients and Q-learning can achieve good performance in challenging dialogue scenarios \citep{fatemi2016policy,su2017sample}.

However, in contrast to  other RL domains, the lack of a common testbed for spoken dialogue has made it difficult to compare different algorithms.
Recent RL advancements
have been largely influenced by the release of benchmarking environments \cite{bellemare2013arcade,duan2016benchmarking} which allow a fair comparison to be made of different RL algorithms operating under similar conditions. 

In the same spirit, based on the recently released PyDial multi-domain SDS tool-kit \cite{ultes2017pydial}, this paper aims to provide a set of testbed environments for developing and evaluating dialogue models. 
To account for the large variability of different scenarios, these environments span different size domains, different user behaviours and different input channel performance. To provide some baselines, the evaluation of a set of the most representative reinforcement learning algorithms for DM is presented. The benchmark and environment implementations are available on-line\footnote{\url{http://www.camdial.org/pydial/benchmarks/}}, allowing for the development, implementation, and evaluation of new algorithms and tasks.

\section{Motivation and related work}
During the last decade, several reinforcement learning algorithms have been applied to the task of dialogue policy optimization \cite{levin1998using,henderson2005hybrid,POMDP_williams,pietquin2011sample,jurcicek2010natural,gasic2014gaussian,su2017sample}. However, the evaluations of these algorithms are hard to compare, mostly because of the lack of a common benchmark environment. In addition, they are usually evaluated in only a few environments, making it hard to assess their potential to generalise to different environments.

In other fields, such as video game playing \cite{bellemare2013arcade,vinyals2017starcraft} and continuous control \cite{duan2016benchmarking}, the release of common benchmarking environments has been a great stimulus to  research in that area, leading to achievements such as human level game playing \cite{mnih2015human} or beating the world champion in the game of Go \cite{silver2016mastering}.

Historically, there has not been a common testbed for the dialogue policy optimisation task. There are several reasons for this. First of all, unlike supervised learning tasks, using a corpus of dialogues to train a RL algorithm can be used only in a bootstrapping phase. 
However, a corpus can not be used to evaluate the final outcome of a dialogue\footnote{It can be used to evaluate the per-turn responses though \cite{fb_n2n}.}, because the learning of an RL agent involves sequential observation and feedback generated from its operating world. This feedback, is conditioned on the action of the agent itself. Therefore, two different policies will generate two different sequences of observations. Training and testing policies directly interacting with real users has been proposed \cite{milica_real_users}. However, system complexity, time and high cost make this approach infeasible for a large part of the research community. In addition, it can be very hard to control for extraneous factors that can modify the behaviour of users such as mood or tiredness, making a fair assessment very difficult. 

To cope with these problems, simulated users \cite{userSim,pietquin2006probabilistic,asri2016sequence,keizer2010simuser} (and simulated input processing channels \citep{pietquin2002asr,thomson2012n}) have been proposed. These models approximate the behaviour of  real users along with input channel noise  introduced by ASR or NLU errors. However, the development of the processing modules needed to create a simulated dialogue environment requires a lot of effort. Even though some simulated environments are publicly available \cite{williams2010demonstration,li2016user, lison2016opendial}, they cover very small dialogue domains and the lack of consistency across them prohibits  wide-scale testing.

The need for a common testbed for the dialogue task is a known issue in the dialogue community, with initiatives such as the Dialogue State Tracking Challenges (DSTC) $1$ to $5$ being the most prominent one \cite{williams2013dialog}. These challenges were possible thanks to a clear evaluation metric. Recently, the BABI dialogue tasks \cite{fb_n2n,li2016dialogue} and the DSTC6, (renamed to Dialogue Systems Technology Challenge), aim to create a testbed for end-to-end text based dialogue management. However, these tasks are focused either on end-to-end supervised learning or in RL based question answering tasks, where the reward signal is delayed only a few steps in time. 

\section{Dialogue management through reinforcement learning}
Dialogue management can be cast as a continuous MDP \citep{young2013pomdp} composed of a continuous multivariate belief state space $\mathcal{B}$, a finite set of actions $\mathcal{A}$ and a reward function $R(b_t,a_t)$. The belief state $b$ is a probability distribution over all possible (discrete) states. At a given time $t$, the agent (policy) observes the belief state $b_t \in \mathcal{B}$ and executes an action $a_t\in \mathcal{A}$. The agent then receives a reward $r_t \in \mathbb{R}$ drawn from $R(b_t,a_t)$. 

The policy $\pi$ is defined as a function $\pi : \mathcal{B} \times \mathcal{A} \rightarrow [0,1]$ that with probability $\pi(b,a)$ takes an action $a$ in a state $b$. For any policy $\pi$ and $b \in \mathcal{B}$, the value function $V^{\pi}$ corresponding to $\pi$ is defined as:
\begin{equation} \label{eq:value_fun}
\begin{aligned}
V^{\pi}(b)& =  \mathbb{E}\{ r_{t} + \gamma r_{t+1} + ... ~~ | b_t=b, \pi \},
\end{aligned}
\end{equation}
where $\gamma$, $0 \leq \gamma \leq 1$, is a discount factor and $r_t$ is one-step reward. 

The objective of reinforcement learning is to find an optimal policy $\pi^{*}$, i.e. a policy that maximizes the value function in each belief state. Equivalently, we can estimate the unique optimal value function $V^{*}$ which corresponds to an optimal policy.
In both cases, the goal is to find an optimal policy $\pi^*$ that maximises the discounted total return 
\begin{equation}\label{eq:reward}
R = \sum_{t=0}^{T-1} \gamma^t r_t(b_t, a_t)
\end{equation}
 over a dialogue with $T$ turns, where $r_t(b_t, a_t)$ is the reward when taking action $a_t$ in dialogue state $b_t$ at turn $t$ and $\gamma$ is the discount factor. 
 
\subsection{Task oriented dialogue management RL environment}
In the RL framework, the environment encompasses every part of the system which is outside the control of the agent. In a modular SDS, the RL environment is every part of the system except the policy itself.
In most classical approaches, the policy module acts as the agent and the rest of the modules constitute the environment (Figure \ref{fig:rl_env}). However, various ways have been proposed to train the policy jointly with other modules using RL.
For example,  the state tracker and the policy can be trained jointly \cite{zhao2016towards, williams2017hybrid}. Other approaches train the policy and the NLG module jointly \cite{wen2017latent}, learn to query the database and the policy together \cite{yang2017end} or train all the models of a (text based) system jointly \cite{fb_n2n}. In this paper, we focus on the classical approach where only the policy is optimised through reinforcement learning.

There are other design features that have impact on the environment. For example, to reduce the action space of the MDP, the full set of actions can be clustered as summary actions \cite{young2013pomdp,thomson2013statistical,williams2008masks}.
In addition, it is often desirable for SDSs to constraint the set of actions the system can take at each turn (e.g. avoid attempting to book a hotel before the dates have been specified). This is usually done by defining a set of action masks -- i.e. heuristics which reduce the number of actions the MDP can take in each dialogue state \cite{williams2008masks,thomson2013statistical,gavsic2009masks}. The use of action masks also speeds up  learning. However, these heuristics must be carefully defined by the system designer, since a poor design of summary actions or masks can lead to suboptimal policies. 

In addition, the domain (specified by the ontology) determines the state space size (input) and action set size (output) of the MDP, as well as influencing several other modules. See appendix \ref{apx:dial_flow} for a schematic example of the summary action mapping, action mask definition and slot based ontology. 

In summary, the dialogue environment has several sources of variability - domain, user behaviour, input channel (i.e. the semantic error rate, N-bests and confidence score distributions, state tracker behaviour), output channel, action masks, summary actions or database access mechanism. A robust dialogue policy should be able to generalise to all of these conditions.

\subsection{Benchmarked algorithms}
In this section, the algorithms used for benchmarking are described. The detailed explanations of the methods and how they are adapted to dialogue management can be found in \cite{gasic2014gaussian,su2017sample}. In general, all algorithms can be divided in two classes: value-based and policy gradient methods.

\textbf{Value-based methods}.~
Value-based methods usually try to estimate a $Q$-value function approximation given a belief state $b$ and an action $a$ with the form:
\begin{equation}
\begin{aligned}
Q^{\pi}(b, a)& = \mathbb{E}_{\pi} \{ r_{t} + \gamma r_{t+1} + ... ~~ | b_t=b, a_t=a\}
\end{aligned}
\end{equation}
where $r_t$ is a one-step reward at time $t$. the policy can be then defined greedily as the action that maximizes $Q^*(b,a)$

\textbf{Policy gradient methods}. ~Value-based models often suffer from divergence problems when using function approximation. This happens because they optimize in value space while following a greedy policy. 
Therefore a slight change in the value function estimate can lead to a large change in the policy space \cite{sutton1999policy}. However, we can directly parametrize a policy $\pi_\theta(a| b)$ and then adjust the parameters to maximize the expected reward (\ref{eq:reward}):
\begin{equation}\label{eq:exp_reward}
\mathbb{E}_{b_0, a_0, ...} \left[ \sum_{t=0}^{T-1} \gamma^t r_t(b_t, a_t) \right]
\end{equation}
where the expectation is taken with respect to all possible dialogue trajectories that start in some initial belief state $b_0$.

To provide some baselines, we investigate a set of representative parametric algorithms, namely deep reinforcement learning algorithms - DQN \citep{mnih2013playing}, A2C \citep{fatemi2016policy} and episodic Natural Actor-Critic (eNAC) \citep{su2017sample} models and compare them to a non-parametric algorithm, GP-SARSA \citep{gasic2014gaussian}. Table \ref{tab:algorithms} presents main characteristics of the four algorithms.

\begin{table}[htbp]
\begin{center}
\resizebox{0.75\columnwidth}{!}{%
\begin{threeparttable}
\begin{tabular}{l|cccc}
 &\multicolumn{1}{c}{GPSARSA} & \multicolumn{1}{c}{DQN}&\multicolumn{1}{c}{A2C} & \multicolumn{1}{c}{eNAC} \\  \hline
\multirow{ 2}{*}{Model type} & non-parametric& parametric & parametric & parametric \\
		 & value-based & value-based  & policy-based & policy-based \\
Value function & \checkmark & \checkmark & \checkmark &\\ 
Policy function &  &  & \checkmark & \checkmark \\ 
Experience replay &  & \checkmark & \checkmark & \checkmark\\ 
Trained by backpropagation &  & \checkmark & \checkmark &  \\ 
Computational complexity & cubic\tnote{*} & linear & linear & linear \\ 
\end{tabular}
  \begin{tablenotes}
  \item[*]In the size of a set of representative points \citep{gasic2014gaussian}.
  \end{tablenotes}
  \end{threeparttable}
  }
    \vspace*{0.1cm}
\caption{General overview of the baseline algorithms analyzed in this benchmark.}
\label{tab:algorithms}
\end{center}
\vspace*{-0.7cm}
\end{table}
\vspace*{-0.1cm}
\subsection{PyDial}
\vspace*{-0.2cm}
PyDial \cite{ultes2017pydial} is an open-source statistical spoken dialogue system toolkit which provides domain-independent implementations of all the dialogue system modules shown in Figure \ref{fig:rl_env}, as well as simulated users and simulated error models. Therefore, this toolkit has the potential to create a set of benchmark environments to compare different RL algorithms in the same conditions. The main focus of PyDial is task-oriented dialogue where the user has to find a matching entity based on a number of constraints. For example the system needs to provide a user with a description of a laptop in a store that meets specific user requirements. In this work, PyDial is used to define different environments, and the configuration files which specify these environments are provided with the paper.

\vspace*{-0.3cm}
\section{Benchmarking tasks}
\vspace*{-0.3cm}
RL-based DM research is typically evaluated on only a single or a very small set of environments.
Such tests do not reveal much about the capability of the algorithms to generalise to different settings, and may be prone to overfitting to specific cases. To test the capability of algorithms in different environments, a set of tasks has been defined that spans a wide range of environments across a number of dimensions:

\textbf{Domain}.  The first dimension of variability between environments is the application domain. Three ontologies with databases of differing sizes are defined, representing information seeking tasks for restaurants in Cambridge and San Francisco and a generic shopping task for laptops \citep{Mrksic15}. These are slot-based ontologies \citep{Henderson2014b}, where the dialogue state is factorised into slots (see appendix \ref{apx:dial_flow} for an factorised state space example). Table \ref{tab:domains} provides a summary of the characteristics of each domain.

\begin{table}[htbp]
\begin{center}
\resizebox{0.7\textwidth}{!}{%
\begin{tabular}{l| c c c c}
Domain & Code & \multicolumn{1}{c}{\# constraint slots} & \multicolumn{1}{c}{\# requests}& \multicolumn{1}{c}{\# values} \\ \hline 
Cambridge Restaurants & CR & 3 & 9 & 268\\ 
San Francisco Restaurants & SFR & 6 & 11 & 636 \\ 
Laptops & LAP & 11 & 21 & 257\\ 
\end{tabular}}
\vspace{1em}
\caption{Description of the domains. The third column represents the number of database search constraints that the user can define, the fourth the number of information slots the user can request from a given database entry and the fifth the sum of the number of values of each requestable slot.}
\label{tab:domains}
\end{center}
\vspace*{-0.6cm}
\end{table}

\textbf{Input error}.  The second dimension of variability comes from the ASR and NLU channel simulation modelling. In PyDial, this is modelled at a semantic level whereby  the true user act is corrupted by noise to generate an N-best-list with associated confidence scores \citep{thomson2012n}. The Semantic Error Rate (SER) is set to three different values, simulating different noise levels in the speech understanding input channel. 

\textbf{User model}. The third dimension of variability comes from the user model. Even if the parameters of the model are sampled at the beginning of each dialogue, the distribution from where these parameters are sampled can be different. In addition to the \textit{Standard} parameter sampling distribution, we define an \textit{Unfriendly} distribution, where the users barely provide any extra information to the system.

\textbf{Masking mechanism}. Finally, in order to test the learning capability of the algorithms, the action masking mechanism provided in PyDial is disabled in two of the tasks. 

In total, $6$ \textit{user model/error model/action mask} environments are defined, representing environments with $0\%, 15\%$, and $30\%$ SER, with the masks deactivated in two of them. Moreover, the parameters of the user behaviour model \citep{schatzmann2007agenda} are sampled at the beginning of each dialogue, simulating the situation that every interaction is conducted with a unique user. Two parameter sampling distributions are defined, \textit{standard} and \textit{unfriendly}. Thus, as summarised in   Table \ref{tab:tasks}, a total of $6*3=18$ different tasks are defined for evaluating each algorithm.

\begin{table}[htbp]
\resizebox{\textwidth}{!}{%
\begin{tabular}{l|c c c|c c c|c c c|c c c|c c c|c c c}
 & \multicolumn{ 3}{c|}{Env. 1} & \multicolumn{ 3}{c|}{Env. 2} & \multicolumn{ 3}{c|}{Env. 3} & \multicolumn{ 3}{c|}{Env. 4} & \multicolumn{ 3}{c}{Env. 5} & \multicolumn{ 3}{c}{Env. 6} \\ 
 \textit{task}& T1.1 & T1.2 & T1.3 & T2.1 & T2.2 & T2.3 & T3.1 & T3.2 & T3.3 & T4.1 & T4.2 & T4.3 & T5.1 & T5.2 & T5.3 & T6.1 & T6.2 & T6.3 \\ \hline \hline
Domain & CR & SFR & LAP & CR & SFR & LAP & CR & SFR & LAP & CR & SFR & LAP & CR & SFR & LAP & CR & SFR & LAP \\
SER & \multicolumn{ 3}{c|}{0\%} & \multicolumn{ 3}{c|}{0\%} & \multicolumn{ 3}{c|}{15\%} & \multicolumn{ 3}{c|}{15\%} & \multicolumn{ 3}{c|}{15\%}& \multicolumn{ 3}{c}{30\%} \\ 
Masks & \multicolumn{ 3}{c|}{On} & \multicolumn{ 3}{c|}{Off} & \multicolumn{ 3}{c|}{On} & \multicolumn{ 3}{c|}{Off} & \multicolumn{ 3}{c|}{On}& \multicolumn{ 3}{c}{On} \\
User & \multicolumn{ 3}{c|}{Standard} & \multicolumn{ 3}{c|}{Standard} & \multicolumn{ 3}{c|}{Standard} & \multicolumn{ 3}{c|}{Standard} & \multicolumn{ 3}{c|}{Unfriendly}& \multicolumn{ 3}{c}{Standard} \\
\end{tabular}}
\vspace{1em}
\caption{The set of benchmarking tasks. Each \textit{user model/error model/action mask} environment is evaluated in three different domains.}
\label{tab:tasks}
\vspace*{-0.6cm}
\end{table}

\vspace*{-0.1cm}
\section{Experimental Setup}
\vspace*{-0.3cm}
In this section, the experimental setup used to run the benchmarking tasks is explained.
\vspace*{-0.2cm}
\subsection{Simulated user and input channel}
\vspace*{-0.2cm}
The user behaviour is modelled by an agenda-based simulator which provides semantic-level interactions \cite{schatzmann2007agenda}. The actions taken during each dialogue are conditioned by parameters sampled from a user model. These are re-sampled at the beginning of each dialogue to ensure a unique profile for every dialogue. The user model has $26$ parameters (e.g. probabilities determining the frequency of repetitions and confirmations), and the range over which the parameters are sampled is provided by a PyDial configuration file. The semantic error rate introduced by the noisy speech channel is simulated through an error model \citep{casanueva2014adaptive,thomson2012n} with parameters learned from real NLU data\footnote{In order to ensure variability, the parameters of environments with different error rate are trained from different data - i.e. the environments are grouped in (1, 2), (3, 4, 5) and (6), each group having different parameters.} \cite{mrkvsic2016neural}. This model has $41$ parameters (e.g. specifying the variability of confidence scores in the input N-best-list).  

All tasks use the same rule-based dialogue state tracker. 
It factorises the dialogue state distribution into the $|S|$ different slots defined by the ontology, plus several general slots 
which track dialogue meta-data, e.g. whether or not the user has been presented with some entity.
Each slot has $|V_s|$ values also defined by the ontology. For a more detailed description of the state tracker refer to \cite{Henderson2014b}. 
\vspace*{-0.2cm}
\subsection{Summary actions and action masks}
\vspace*{-0.2cm}
The MDP action set is defined as a set of summary actions \citep{thomson2013statistical,young2010hidden}. This set consists of $5$ slot independent actions (\textit{inform by constraints}, \textit{inform requested}, \textit{inform alternatives}, \textit{bye} and \textit{request more}) and $3$ slot dependent actions (\textit{request}, \textit{confirm} and \textit{select}), making a total of $5+3*|S|$ actions where $|S|$ is the number of slots requestable slots (see Tab. \ref{tab:domains}). The mapping between summary and master actions is based on simple heuristics dependent on the belief state (e.g, inform a venue matching the top values of each slot, confirm the top value of a slot, etc.)

In the case of action masks, similar heuristics are used. For slot dependent actions, these heuristics depend on the distribution of the values of that slot (e.g. \textit{confirm foodtype} is masked if all the probability mass of \textit{foodtype} is in the "none" value). For the slot independent actions, the masks depend on the general \textit{method} slot, which tracks the way the user is conducting the database search. The masks of the slot independent actions are dependent on the value of this slot (e.g. \textit{inform by constraints} is only unmasked if the top value of the \textit{method} slot is \textit{byconstraints}).
\vspace*{-0.2cm}
\subsection{Model hyperparameters}
\vspace*{-0.2cm}
GP-SARSA uses a \textit{linear kernel} for the state space and a \textit{delta kernel} for the action space. The \textit{scale}, responsible for the degree of exploration, is set $3$. The remaining parameters are set as in \cite{gasic2014gaussian}. Futher improvements in overall performance can be obtained with a Gaussian kernel with optimized hyperparameters~\cite{chen2015hyper}, however this was not explored here.

Unlike GPSARSA, the trade-off between exploration and exploitation is not handled automatically in deep-RL models, being dependent on the number of training dialogues. The exploration schedule  is often a critical factor in obtaining good learning performance. 
The $\epsilon$-greedy policy used here follows a linear scheduling starting from $\epsilon$ and then annealed to $0.05$ after 4000 dialogues, where the optimal initial value for $\epsilon$ was found by grid search over values $0.9$, $0.5$ and $0.3$. 

All deep-RL policy models are composed of $2$ hidden feedforward layers. As the objective of the paper is to see how these models generalise across environments, the hyper-parameters of all models across all the tasks are kept the same. The hyperparameters are set as in \cite{su2017sample}, with the exception of the size of the hidden layers and the initial $\epsilon$, which are tuned by grid search. Table \ref{tab:architecture} presents the hyperparameters of the best models across each domain for all deep-RL algorithms, selected through a grid search over $9$ combinations of hyperparameters. The Adam optimiser was used to train all the deep-RL models, with an initial learning rate of $0.001$ \cite{kingma2014adam}.

For a more detailed description, the hyperparameters of every implemented model are specified in the PyDial configuration files provided for each task.
\vspace*{-0.2cm}
\subsection{Handcrafted policy}
\vspace*{-0.2cm}
In addition to the RL algorithms described in Table \ref{tab:algorithms}, the performance of a classic handcrafted policy interacting with each environment is also evaluated. The actions taken by this policy are based on carefully designed heuristics, dependent on the belief state \citep{thomson2013statistical}. 
\vspace*{-0.2cm}
\subsection{Reward function and performance metrics}
\vspace*{-0.2cm}
The maximum dialogue length was set to $25$ turns and the discount factor $\gamma$ was $0.99$. The metrics presented in next section are the average success rate and average reward for each evaluated policy model. Success rate is defined as the percentage of dialogues which are completed successfully -- i.e. whether the dialogue manager is able to fulfill the user goal or not. Final reward is defined as $20*\mathds{1}(\mathcal{D})- T$, where $\mathds{1}(\mathcal{D})$ is the success indicator and $T$ is the dialogue length in turns.
\vspace*{-0.2cm}
\section{Results and discussion}\label{sec:results}
\vspace*{-0.3cm}

\begin{table}[htbp]
\resizebox{0.99\textwidth}{!}{%
\begin{tabular}{llrrrrrrrrrr}
 &  & \multicolumn{2}{c}{GP-Sarsa}  & \multicolumn{2}{c}{DQN} & \multicolumn{2}{c}{A2C}  & \multicolumn{2}{c}{eNAC} & \multicolumn{2}{c}{Handcrafted} \\ 
\multicolumn{2}{c}{\textit{Task}}  & \multicolumn{1}{c}{Suc.} & \multicolumn{1}{c}{Rew.} & \multicolumn{1}{c}{Suc.} & \multicolumn{1}{c}{Rew.} & \multicolumn{1}{c}{Suc.} & \multicolumn{1}{c}{Rew.} & \multicolumn{1}{c}{Suc.} & \multicolumn{1}{c}{Rew.} & \multicolumn{1}{c}{Suc.} & \multicolumn{1}{c}{Rew.} \\
\hline \hline
\multirow{3}{*}{\rotatebox[origin=c]{90}{Env. 1}} 
 & CR & 99.4\% & \textbf{13.5} & 93.9\% & 12.7 & 89.3\% & 11.6 & 94.8\% & 12.4 & 100.0\%&14.0 \\ 
 & SFR & 96.1\% & 11.4 & 65.0\% & 5.9 & 58.3\% & 4.0 & 94.0\% & \textbf{11.7} & 98.2\%&12.4 \\ 
 & LAP & 89.1\% & 9.4 & 70.1\% & 6.9 & 57.1\% & 3.5 & 91.4\% & \textbf{10.5} & 97.0\%&11.7 \\ 
\hline
\multirow{3}{*}{\rotatebox[origin=c]{90}{Env. 2}}
  & CR & 96.8\% & \textbf{12.2} & 91.9\% & 12.0 & 75.5\% & 7.0 & 83.6\% & 9.0 & 100.0\%&14.0 \\ 
 & SFR & 91.9\% & \textbf{9.6} & 84.3\% & 9.2 & 45.5\% & -0.3 & 65.6\% & 3.7 & 98.2\%&12.4 \\ 
 & LAP & 82.3\% & \textbf{7.3} & 74.5\% & 6.6 & 26.8\% & -5.0 & 55.1\% & 1.5 & 97.0\%&11.7 \\
\hline
\multirow{3}{*}{\rotatebox[origin=c]{90}{Env. 3}}
 & CR & 95.1\% & 11.0 & 93.4\% & \textbf{11.9} & 74.6\% & 7.3 & 90.8\% & 11.2 & 96.7\%&11.0 \\
 & SFR & 81.6\% & 6.9 & 60.9\% & 4.0 & 39.1\% & -2.0 & 84.6\% & \textbf{8.6} & 90.9\%&9.0 \\ 
 & LAP & 68.3\% & 4.5 & 61.1\% & 4.3 & 37.0\% & -1.9 & 76.6\% & \textbf{6.7} & 89.6\%&8.7 \\ 
\hline
\multirow{3}{*}{\rotatebox[origin=c]{90}{Env. 4}}
 & CR & 91.5\% & 9.9 & 90.0\% & \textbf{10.7} & 64.7\% & 3.7 & 85.3\% & 9.0 & 96.7\%&11.0 \\ 
 & SFR & 81.6\% & 7.2 & 77.8\% & \textbf{7.7} & 38.8\% & -3.1 & 61.7\% & 2.0 & 90.9\%&9.0 \\ 
 & LAP & 72.7\% & 5.3 & 68.7\% & \textbf{5.5} & 27.3\% & -6.0 & 52.8\% & -0.8 & 89.6\%&8.7 \\  
\hline
\multirow{3}{*}{\rotatebox[origin=c]{90}{Env. 5}}
 & CR & 93.8\% & 9.8 & 90.7\% & 10.3 & 70.1\% & 5.0 & 91.6\% & \textbf{10.5} & 95.9\%&9.7 \\ 
 & SFR & 74.7\% & 3.6 & 62.8\% & 2.9 & 20.2\% & -5.9 & 74.4\% & \textbf{4.5} & 87.7\%&6.4 \\ 
 & LAP & 39.5\% & -1.6 & 45.5\% & 0.0 & 28.9\% & -4.7 & 75.8\% & \textbf{4.1} & 85.1\%&5.5 \\ 
\hline
\multirow{3}{*}{\rotatebox[origin=c]{90}{Env. 6}}
  & CR & 89.6\% & 8.8 & 87.8\% & \textbf{10.0} & 62.3\% & 3.5 & 79.6\% & 8.0 & 89.6\%&9.3 \\ 
 & SFR & 64.2\% & 2.7 & 47.2\% & 0.4 & 27.5\% & -5.1 & 66.7\% & \textbf{3.9} & 79.0\%&6.0 \\ 
 & LAP & 44.9\% & -0.2 & 46.1\% & 1.0 & 32.1\% & -3.8 & 64.6\% & \textbf{3.6} & 76.1\%&5.3 \\ 
\hline
\multirow{3}{*}{\rotatebox[origin=c]{90}{Mean}}
& CR & 94.4\% &	10.9&	91.3\%&	\textbf{11.3}&	72.8\%&	6.4&	87.6\%&	10.0&	96.5\%&	11.5 \\
 & SFR & 81.7\% & \textbf{6.9} & 66.3\% & 5.0 & 38.2\% & -2.1 & 74.5\% & 5.7 & 90.8\%&9.2 \\ 
 & LAP & 66.1\% & 4.1 & 61.0\% & 4.1 & 34.9\% & -3.0 & 69.4\% & \textbf{4.3} & 89.1\%&8.6 \\ 
 & ALL & 80.7\% & \textbf{7.3} & 72.9\% & 6.8 & 48.6\% & 0.4 & 77.2\% & 6.7 & 92.1\%&9.8 \\  
\hline
\end{tabular}
}
\caption{Reward and success rates after $4000$ training dialogues for the five policy models considered in this benchmark. Each row represents one of the $18$ different tasks. The highest reward obtained by a data driven model in each row is highlighted.}
\label{tab:results}
\end{table}

The evaluation results for the $18$ tasks\footnote{As shown in table \ref{tab:tasks}, we refer to the tasks as \textit{task X.Y}, where \textit{X} indicates the \textit{user/error/mask} environment and \textit{Y} the domain. e.g. \textit{Task 2} refers to \textit{env. 2} in the three domains. \textit{Task 2.3} refers to \textit{env. 2} in \textit{LAP}.} are presented in Table \ref{tab:results}. For each task, every model is trained over ten different random seeds and evaluated after $4000$ training dialogues. The models are evaluated over $500$ test dialogues and the results shown are averaged over all $10$ seeds. In addition, evaluation results after $1000$ and $10000$ training dialogues are shown in Appendix \ref{apx:results} and learning curves for task $3$ are shown in Appendix \ref{apx:plots}.

The results clearly show that the domain complexity plays a crucial role on the overall performance.
Value-based methods (GP-SARSA and DQN) achieve the best performance in the CR domain across all six environmental settings. Value-based methods are known to have a higher learning rate. While this might lead to overfitting to the two larger domains (SFR and LAP), in domains with small action and state spaces, a higher learning rate helps to achieve a good policy faster than policy gradient based methods. 
On the other hand, eNAC provides the best performance on the SFR and LAP tasks suggesting that policy-gradient methods scales robustly to larger state and action spaces.

Action masks significantly reduce the size of the action space and thus increase the policy learning rate.
However, for the environments where the action masks are deactivated ($2$ and $4$), the policy-gradient methods learn much slower. In contrast, value-based approaches still maintain reasonable performance, indicating that they are more sample-efficient than policy-based methods. 
However, it is worth noting that DQN is highly unstable, especially with larger domains (see Figure \ref{fig:Ng2}). Thanks to the non-parametric approach, this pattern is not observed with GP-SARSA.
As noted earlier, this is mainly due to optimisation being performed in value space rather than directly in policy space. In addition, after $10K$ dialogues, the performance of eNAC decreases in some environments. This might be because the hyperparameters were optimised for $4K$ dialogues. A more extensive grid search could solve the problem.


The performance of every model drops substantially when noise is introduced to the semantic input. Results from tasks $3$, $4$, $5$ and $6$ show, however, that eNAC is more robust in these partially observed environments and thus degrades less than the other methods.
One  reason for this is that, contrary to other deep-RL methods, the natural gradient points more directly to the desired goal and is less prone to getting stuck on local plateaus,  thereby learning better policies in noisy environments. 

As it could be expected, interacting with the \textit{unfriendly} set of users in \textit{task 5} degrades the performance. However, the performance drop is smaller for eNAC than for the rest of the models. This suggests that this policy has the ability to learn faster how to guide the dialogue when the user is less prone to provide information about his or her goal.

GPSARSA consistently performs well, showing very stable performance and fast learning rate (see Appendix \ref{apx:plots}). Overall, it is the best model across all tasks and domains both in terms of the learning rate and the final performance, followed closely by DQN and eNAC\footnote{Note, however, that the mean results for eNAC are degraded because of the very poor performance in unmasked environments. If this problem could be solved (e.g. by using techniques to increase the sample efficiency\citep{wang2016sample}), this would be the best performing model.}. A2C shows the worse results of all and, contrary to other RL applications, the ability to perform asynchronous learning is less useful because it significantly raises the training costs with real users. It can also be observed that some deep-RL models are prone to overfitting. Furthermore, these algorithms are very sensitive to hyper-parameter values. 

Lastly, it is worth noting that the handcrafted policy model outperforms the RL-based policies in almost all the tasks in the larger domains (SFR and LAP),  showing that RL-based models have difficulties to learn in large state spaces. To mitigate this issue, state space abstraction~\citep{wang2015learning,papangelis2017single} or hierarchical reinforcement learning~\cite{budzianowski2017subdomain} approaches can be used.
\subsection{Cross-tasks evaluation}
\vspace*{-0.2cm}
To further examine the generalisation capabilities of the various algorithms, we performed some cross-task evaluations. We chose three tasks, namely $1$, $3$ and $6$ to test how algorithms trained in a noisy environment perform in a zero noise set-up and vice versa. Table \ref{tab:crosstasks} presents results for GPSARSA, DQN and A2C. For clarity, we omit A2C results since this algorithm performed substantially worse.

Results show that eNAC has the strongest generalisation capabilities, having the best performance in most of the cross-task environments. Value-based models have a good performance when trained with noisy data and tested in clean data, with DQN getting very close performance to eNAC. However, when trained in clean data and tested in noisy data the performance greatly decreases, especially in the larger domains. This decrease in performance is more severe for GPSARSA.




\section{Conclusions and future work}
To our knowledge, this is the first work to present a set of extensive simulated dialogue management environments along with a comparison of several RL algorithms using an open-domain toolkit. The results show that a large amount of improvement is still necessary for data driven models to match the performance of handcrafted policies, especially in larger domains. The environments presented in this paper, however, are still very constrained compared to real world tasks (e.g. Siri, Alexa...). In the future, we  plan to include multi-domain environments (where the rewards are more delayed in time and are thus more challenging environments) and word-level user simulations, which would enable the dialogue managers to be trained in more realistic environments. Also, these environments are implemented in an open domain toolkit, offering the possibility to the research community to add new algorithms and new tasks.

\subsubsection*{Acknowledgments}
This research was partly funded by the
EPSRC grant EP/M018946/1 Open Domain Statistical
Spoken Dialogue Systems. Paweł
Budzianowski  is  supported  by  EPSRC  Council and Toshiba Research Europe Ltd, Cambridge Research  Laboratory.  Pei-Hao  Su  is  supported  by  Cambridge  Trust and  the  Ministry  of  Education,  Taiwan. The benchmark is available on-line at \url{http://www.camdial.org/pydial/benchmarks/}.

\newpage
\bibliography{biblio}
\bibliographystyle{plainnat}

\appendix

\section{Example of the dialogue flow in a modular SDS}\label{apx:dial_flow}
\vspace*{-0.6cm}
\begin{figure}[h]
  \centering
 \includegraphics[width=\textwidth]{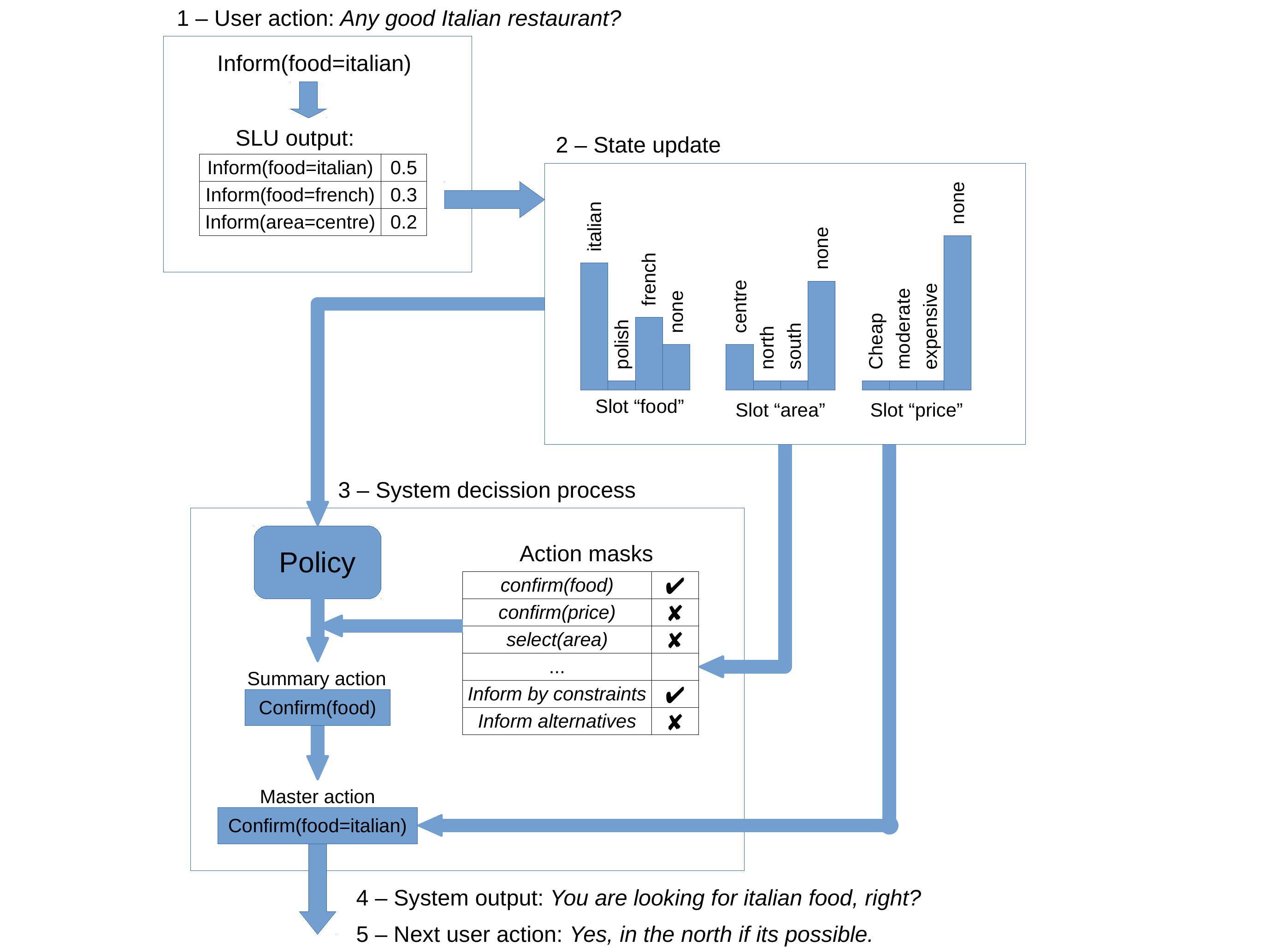}
  \caption{Dialogue flow during a single turn in a modular SDS. The turn begins with the user action (1), which is processed by the input channel and converted into a N-best list with confidence scores. This N-best list is used by the state tracker (2) to update the dialogue (belief) state. The dialogue state is used by several modules of the system decision process (3). First, the action masks are defined based in the dialogue state. Then, the policy choses the optimal summary action based on the dialogue state and the action masks. Finally, the summary action is converted into a master action using heuristics based on the dialogue state. Then, the system outputs the action (usually through an NLG+TTS channel) (4) and the cycle begins again (5).}
  \label{fig:dial_flow}
\end{figure} 

\vspace*{-0.6cm}
\section{Architecture details}\label{apx:arch}
\begin{table}[htbp]
\begin{center}
\begin{tabular}{lccc}
 Model & \multicolumn{1}{l}{Hidden layer 1}&\multicolumn{1}{l}{Hidden layer 2} & \multicolumn{1}{l}{$\epsilon$ starting value} \\ \hline \hline
  DQN & 300 &100 & 0.3 \\ 
  A2C & 200&75 & 0.5 \\ 
  eNAC & 130&50 & 0.3 \\ 
\end{tabular}
\vspace{1.5em}
\caption{Hyperparameters of the deep-RL models}
\label{tab:architecture}
\end{center}
\end{table}
\newpage
\section{Results after 1000 and 10000 training dialogues}\label{apx:results}
\begin{table}[htbp]
\resizebox{0.97\textwidth}{!}{%
\begin{tabular}{llrrrrrrrrrr}
 &  & \multicolumn{2}{c}{GP-Sarsa}  & \multicolumn{2}{c}{DQN} & \multicolumn{2}{c}{A2C}  & \multicolumn{2}{c}{ENAC} & \multicolumn{2}{c}{Handcrafted} \\ 
\multicolumn{2}{c}{\textit{Task}}  & \multicolumn{1}{c}{Suc.} & \multicolumn{1}{c}{Rew.} & \multicolumn{1}{c}{Suc.} & \multicolumn{1}{c}{Rew.} & \multicolumn{1}{c}{Suc.} & \multicolumn{1}{c}{Rew.} & \multicolumn{1}{c}{Suc.} & \multicolumn{1}{c}{Rew.} & \multicolumn{1}{c}{Suc.} & \multicolumn{1}{c}{Rew.} \\
\hline \hline
\multirow{3}{*}{\rotatebox[origin=c]{90}{Env. 1}} 
 & CR & 98.0\% & \textbf{13.0} & 88.6\% & 11.6 & 83.4\% & 10.0 & 93.0\% & 12.2 & 100.0\%&14.0 \\
 & SFR & 91.9\% & \textbf{10.0} & 48.0\% & 2.7 & 46.9\% & 1.7 & 85.8\% & 9.9 & 98.2\%&12.4 \\ 
 & LAP & 78.9\% & 6.7 & 61.9\% & 5.5 & 41.1\% & 0.3 & 84.2\% & \textbf{8.8} & 97.0\%&11.7 \\ 
\hline
\multirow{3}{*}{\rotatebox[origin=c]{90}{Env. 2}}
 & CR & 91.1\% & \textbf{10.8} & 67.6\% & 6.4 & 58.6\% & 4.0 & 78.8\% & 6.6 & 100.0\%&14.0 \\
 & SFR & 82.1\% & \textbf{7.4} & 64.2\% & 5.4 & 39.0\% & -1.1 & 67.5\% & 2.7 & 98.2\%&12.4 \\ 
 & LAP & 68.4\% & 3.1 & 70.8\% & \textbf{5.8} & 31.0\% & -3.6 & 57.8\% & -0.5 & 97.0\%&11.7 \\
\hline
\multirow{3}{*}{\rotatebox[origin=c]{90}{Env. 3}}
 & CR & 91.9\% & \textbf{10.4} & 79.5\% & 9.2 & 66.5\% & 6.0 & 85.7\% & 10.0 & 96.7\%&11.0 \\ 
 & SFR & 76.6\% & 5.5 & 42.4\% & 1.0 & 34.5\% & -2.1 & 73.6\% & \textbf{6.2} & 90.9\%&9.0 \\ 
 & LAP & 65.0\% & 2.8 & 51.9\% & 3.1 & 32.7\% & -2.2 & 71.0\% & \textbf{5.5} & 89.6\%&8.7 \\ 
\hline
\multirow{3}{*}{\rotatebox[origin=c]{90}{Env. 4}}
 & CR & 88.2\% & \textbf{9.3} & 73.5\% & 6.9 & 54.2\% & 2.2 & 73.6\% & 4.4 & 96.7\%&11.0 \\ 
 & SFR & 73.6\% & \textbf{4.9} & 65.9\% & 4.5 & 27.2\% & -3.7 & 60.4\% & 0.8 & 90.9\%&9.0 \\ 
 & LAP & 61.3\% & 0.3 & 53.2\% & \textbf{2.7} & 28.1\% & -3.7 & 46.9\% & -2.9 & 89.6\%&8.7 \\ 
\hline
\multirow{3}{*}{\rotatebox[origin=c]{90}{Env. 5}}
 & CR & 90.2\% & \textbf{9.0} & 60.1\% & 4.1 & 49.0\% & 1.6 & 81.2\% & 8.1 & 95.9\%&9.7 \\ 
 & SFR & 65.3\% & \textbf{1.3} & 32.5\% & -2.0 & 14.0\% & -6.2 & 54.0\% & 0.9 & 87.7\%&6.4 \\ 
 & LAP & 44.9\% & -2.8 & 31.4\% & -1.8 & 17.8\% & -5.5 & 61.3\% & \textbf{1.7} & 85.1\%&5.5 \\ 
\hline
\multirow{3}{*}{\rotatebox[origin=c]{90}{Env. 6}}
 & CR & 84.9\% & \textbf{8.3} & 72.3\% & 6.9 & 50.2\% & 2.1 & 73.6\% & 6.7 & 89.6\%&9.3 \\ 
 & SFR & 59.7\% & 0.7 & 35.6\% & -1.2 & 19.0\% & -5.6 & 55.2\% & \textbf{1.4} & 79.0\%&6.0 \\ 
 & LAP & 52.0\% & -1.5 & 47.5\% & 1.4 & 20.7\% & -5.3 & 56.3\% & \textbf{1.9} & 76.1\%&5.3 \\ 
\hline
\multirow{3}{*}{\rotatebox[origin=c]{90}{Mean}}
 & CR & 90.7\% & \textbf{10.1} & 73.6\% & 7.5 & 60.3\% & 4.3 & 81.0\% & 8.0 & 96.5\%&11.5 \\ 
 & SFR & 74.9\% & \textbf{5.0} & 48.1\% & 1.7 & 30.1\% & -2.8 & 66.1\% & 3.6 & 90.8\%&9.2 \\ 
 & LAP & 61.7\% & 1.4 & 52.8\% & \textbf{2.8} & 28.6\% & -3.3 & 62.9\% & 2.4 & 89.1\%&8.6 \\ 
 & ALL & 75.8\% & \textbf{5.5} & 58.2\% & 4.0 & 39.7\% & -0.6 & 70.0\% & 4.7 & 92.1\%&9.8 \\ 
 \hline
\end{tabular}
}
\caption{Reward and success rates after $1000$ training dialogues.}
\label{tab:results1K}
\end{table}

\begin{table}[htbp]
\resizebox{0.97\textwidth}{!}{%
\begin{tabular}{llrrrrrrrrrr}
 &  & \multicolumn{2}{c}{GP-Sarsa}  & \multicolumn{2}{c}{DQN} & \multicolumn{2}{c}{A2C}  & \multicolumn{2}{c}{ENAC} & \multicolumn{2}{c}{Handcrafted} \\ 
\multicolumn{2}{c}{\textit{Task}}  & \multicolumn{1}{c}{Suc.} & \multicolumn{1}{c}{Rew.} & \multicolumn{1}{c}{Suc.} & \multicolumn{1}{c}{Rew.} & \multicolumn{1}{c}{Suc.} & \multicolumn{1}{c}{Rew.} & \multicolumn{1}{c}{Suc.} & \multicolumn{1}{c}{Rew.} & \multicolumn{1}{c}{Suc.} & \multicolumn{1}{c}{Rew.} \\
\hline \hline
\multirow{3}{*}{\rotatebox[origin=c]{90}{Env. 1}} 
&CR& 99.4\% & \textbf{13.5} & 92.5\% & 12.4 & 86.3\% & 10.5 & 85.3\% & 10.5 & 100.0\%&14.0 \\ 
&SFR&97.3\% & 11.7 & 79.5\% & 8.7 & 65.4\% & 5.4 & 97.0\% & \textbf{12.3} & 98.2\%&12.4 \\ 
&LAP&90.3\% & 9.7 & 72.9\% & 7.3 & 56.0\% & 3.5 & 92.1\% & \textbf{11.0} & 97.0\%&11.7 \\ 
\hline
\multirow{3}{*}{\rotatebox[origin=c]{90}{Env. 2}}
&CR& 97.9\% & 12.4 & 96.1\% & \textbf{12.7} & 66.3\% & 4.4 & 49.4\% & 2.3 & 100.0\%&14.0 \\ 
&SFR&95.4\% & \textbf{10.1} & 84.2\% & 9.7 & 32.9\% & -3.3 & 59.0\% & 3.0 & 98.2\%&12.4 \\ 
&LAP&87.5\% & 8.4 & 83.9\% & \textbf{9.1} & 22.2\% & -6.0 & 42.7\% & -0.2 & 97.0\%&11.7 \\ 
\hline
\multirow{3}{*}{\rotatebox[origin=c]{90}{Env. 3}}
&CR& 95.8\% & 10.9 & 94.7\% & \textbf{12.2} & 81.5\% & 8.4 & 76.0\% & 8.2 & 96.7\%&11.0 \\ 
&SFR&81.2\% & 6.5 & 73.1\% & 6.2 & 37.8\% & -2.9 & 84.1\% & \textbf{8.6} & 90.9\%&9.0 \\ 
&LAP&64.3\% & 3.9 & 69.2\% & 5.6 & 48.5\% & -0.3 & 73.3\% & \textbf{6.5} & 89.6\%&8.7 \\ 
\hline
\multirow{3}{*}{\rotatebox[origin=c]{90}{Env. 4}}
&CR& 92.6\% & 10.0 & 91.9\% & \textbf{11.1} & 61.9\% & 2.3 & 51.6\% & 2.4 & 96.7\%&11.0 \\ 
&SFR&81.0\% & 6.9 & 81.1\% & \textbf{8.2} & 34.1\% & -4.9 & 28.0\% & -3.5 & 90.9\%&9.0 \\ 
&LAP&74.0\% & \textbf{5.8} & 69.3\% & 5.6 & 25.2\% & -7.3 & 35.2\% & -1.7 & 89.6\%&8.7 \\ 
\hline
\multirow{3}{*}{\rotatebox[origin=c]{90}{Env. 5}}
&CR& 91.7\% & 8.8 & 92.6\% & \textbf{10.5} & 67.8\% & 3.9 & 78.9\% & 7.9 & 95.9\%&9.7 \\ 
&SFR&68.6\% & 2.7 & 72.8\% & 4.5 & 23.8\% & -6.3 & 82.3\% & \textbf{6.5} & 87.7\%&6.4 \\ 
&LAP&36.9\% & -1.4 & 53.3\% & 0.7 & 24.7\% & -5.6 & 72.8\% & \textbf{3.8} & 85.1\%&5.5 \\ 
\hline
\multirow{3}{*}{\rotatebox[origin=c]{90}{Env. 6}}
&CR&89.6\% & 8.6 & 88.3\% & \textbf{9.9} & 62.3\% & 2.8 & 57.8\% & 3.9 & 89.6\%&9.3 \\ 
&SFR&54.8\% & 1.3 & 64.8\% & \textbf{3.6} & 23.5\% & -6.3 & 61.1\% & 3.1 & 79.0\%&6.0 \\ 
&LAP&45.6\% & 0.3 & 52.1\% & 1.7 & 25.3\% & -5.6 & 61.2\% & \textbf{3.2} & 76.1\%&5.3 \\  
\hline
\multirow{3}{*}{\rotatebox[origin=c]{90}{Mean}}
&CR&94.5\% & \textbf{10.7} & 92.7\% & 11.5 & 71.0\% & 5.4 & 66.5\% & 5.9 & 96.5\%&11.5 \\ 
&SFR&79.7\% & 6.5 & 75.9\% & \textbf{6.8} & 36.2\% & -3.1 & 68.6\% & 5.0 & 90.8\%&9.2 \\ 
&LAP&66.4\% & 4.4 & 66.8\% & \textbf{5.0} & 33.7\% & -3.6 & 62.9\% & 3.8 & 89.1\%&8.6 \\
&ALL&80.2\% & 7.2 & 78.5\% & \textbf{7.8} & 47.0\% & -0.4 & 66.0\% & 4.9 & 92.1\%&9.8 \\ 
\hline
\end{tabular}
}
\caption{Reward and success rates after $10000$ training dialogues.}
\label{restable}
\end{table}
\newpage
\section{Learning curves for task 3}\label{apx:plots}
\begin{figure}[!h]
\centering
\begin{subfigure}[a]{.85\textwidth}
   \includegraphics[width=1\linewidth]{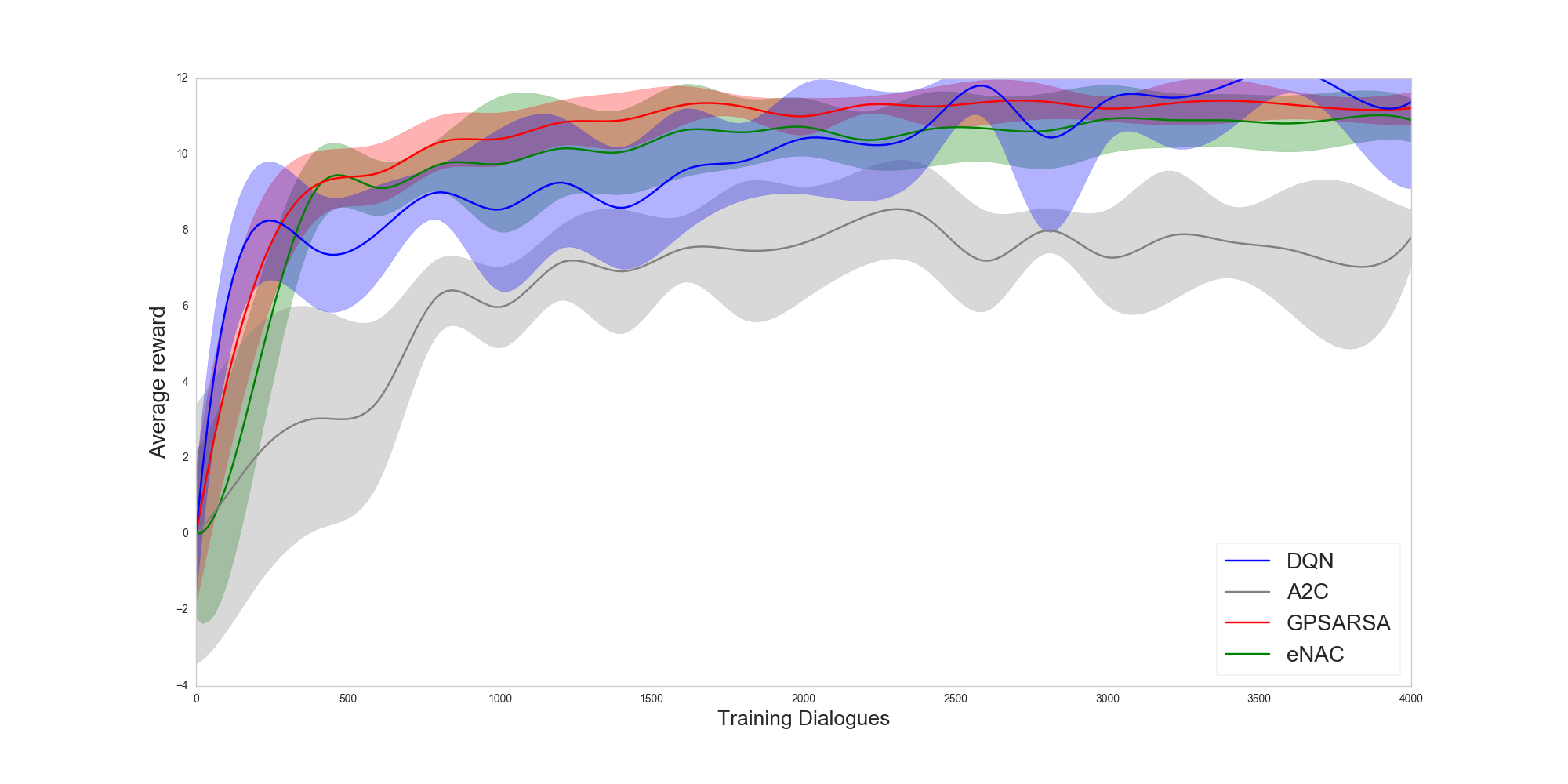}
   \caption{Cambridge Restaurants}
   \label{fig:Ng1} 
\end{subfigure}

\begin{subfigure}[b]{.85\textwidth}
   \includegraphics[width=1\linewidth]{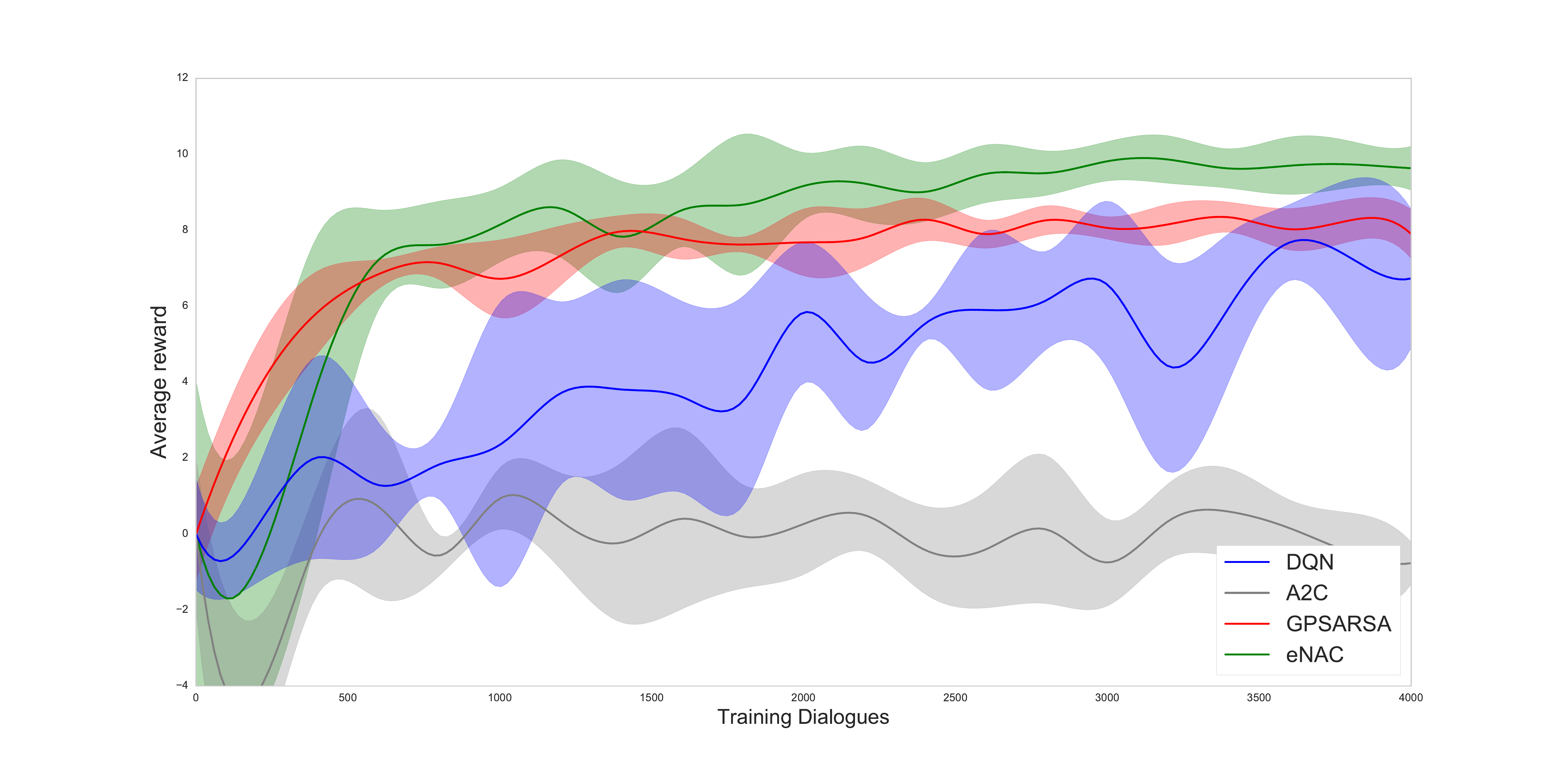}
   \caption{San Francisco Restaurants}
   \label{fig:Ng2}
\end{subfigure}

\begin{subfigure}[c]{.85\textwidth}
   \includegraphics[width=1\linewidth]{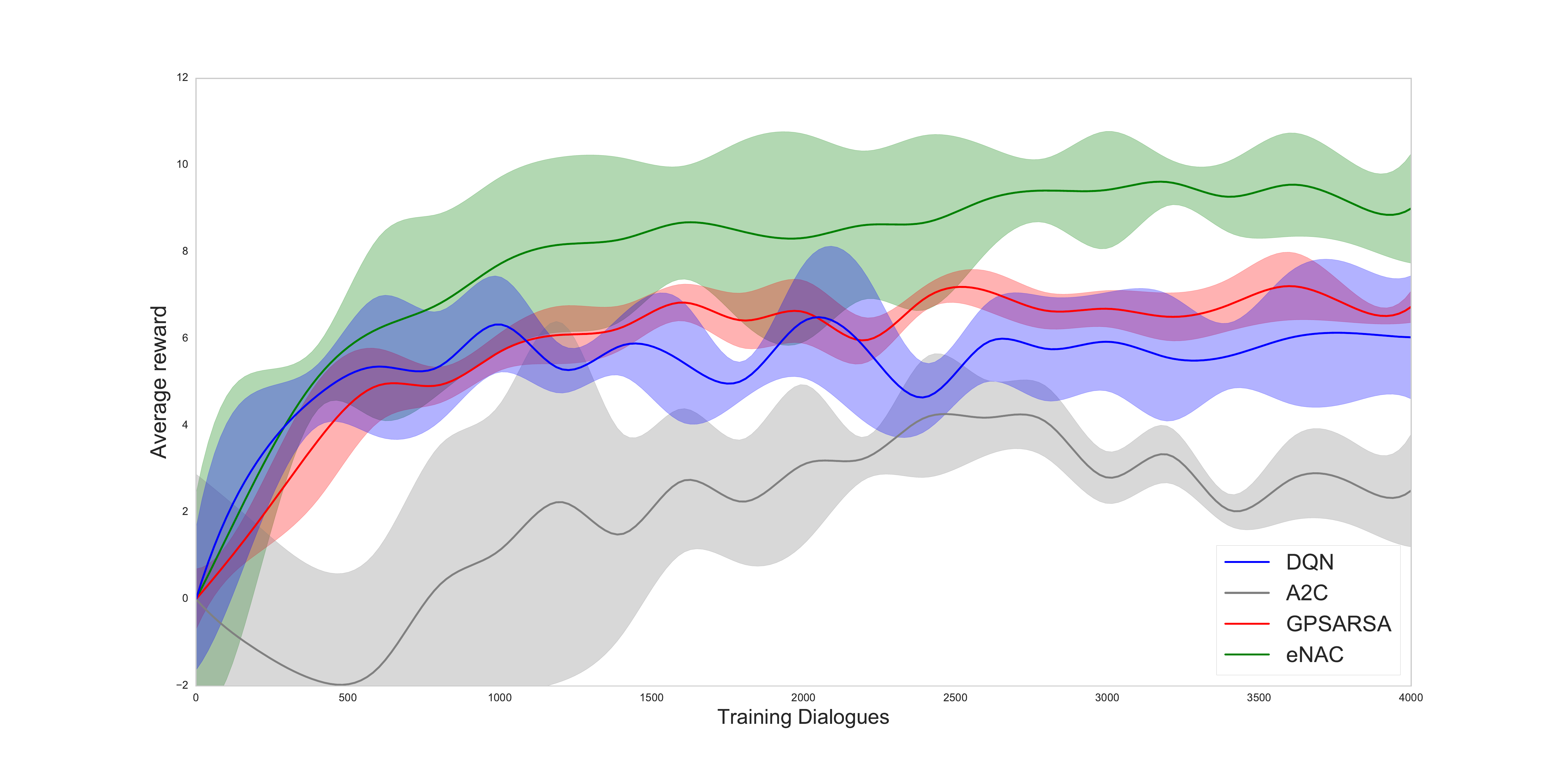}
   \caption{Laptops}
   \label{fig:Ng3}
\end{subfigure}

\caption[Two numerical solutions]{Performance of the benchmarked algorithms as a function of the number of dialogues for three different domains; the shaded area depicts the mean $\pm$ the standard deviation over ten
different random seeds.}
\label{fig:curves}
\end{figure}

\newpage

\section{Cross task experiments}

\begin{table}[htbp]
\begin{tabular}{c|l||lll|lll|lll|}
\multicolumn{1}{l}{} & Evaluation & \multicolumn{3}{c|}{Env. 1} & \multicolumn{3}{c|}{Env. 3} & \multicolumn{3}{c|}{Env. 6}  \\ \hline 
Training & Model/Domain & CR & SFR & LAP & CR & SFR & LAP & CR & SFR&LAP \\ \hline \hline
 \multirow{3}{*}{Env. 1} & GP-SARSA &  & &  & 0.6 & -6.8 & -5.9 & -4.3 & -12.3&-11.0 \\ 
 & DQN &  & &  & 8.0 & 0.2 & 3.4 & 4.6 & -1.8&0.8 \\ 
 & ENAC &  & &  & \textbf{9.7} & \textbf{8.0} & \textbf{7.3} & \textbf{7.0} & \textbf{4.3}&\textbf{4.1} \\ \hline
\multirow{3}{*}{Env. 3} & GP-SARSA & 9.5 & 9.9 & 6.1 &  & && 7.0 & 0.7&-2.7 \\ 
 & DQN & 13.0 & 7.7 & 6.5 &  & & &\textbf{9.1} & 1.9&1.6 \\ 
 & ENAC & \textbf{13.1} & \textbf{10.9} & \textbf{10.3} & &  &  & 7.7 & \textbf{4.7}&\textbf{3.9} \\ \hline
\multirow{3}{*}{Env. 6} & GP-SARSA & 11.9 & 9.1 & 5.1 & 10.6 & 6.1 & 2.6 & & & \\ 
 & DQN & \textbf{13.8} & 6.3 & 6.0 & \textbf{12.0} & 3.7 & 3.7 &  & & \\ 
 & ENAC & 13.2 & \textbf{9.7} & \textbf{10.2} & 10.9 & \textbf{7.2} & \textbf{7.0} &  & & \\ \hline
\end{tabular}
\vspace{1.5em}
\caption{Reward obtained by the three best performing algorithms in cross-tasks evaluation. The (wide) rows represent the environment in which the model is trained and the (wide) columns the environment where its evaluated. The (thin) rows represent the model and the (thin) columns the domain where the model is trained and tested. The reward for the best performing model in each cross-environment setup and domain combination is highlighted.}
\label{tab:crosstasks}
\end{table}

\end{document}